\documentclass[fleqn,10pt]{wlscirep}
\usepackage[utf8]{inputenc}
\usepackage[T1]{fontenc}

\title{Deep learning algorithms out-perform veterinary pathologists in detecting the mitotically most active tumor region}

\author[1,*]{Marc~Aubreville}
\author[2]{Christof~A.~Bertram}
\author[1]{Christian~Marzahl}
\author[3]{Corinne~Gurtner}
\author[3]{Martina~Dettwiler}
\author[4]{Anja~Schmidt}
\author[2]{Florian~Bartenschlager}
\author[2]{Sophie~Merz}
\author[2]{Marco~Fragoso}
\author[2]{Olivia~Kershaw}
\author[2]{Robert~Klopfleisch}
\author[1]{Andreas~Maier}

\affil[1]{Pattern Recognition Lab, Computer Science, Friedrich-Alexander-Universit{\"a}t Erlangen-N{\"u}rnberg, Erlangen, Germany}
\affil[2]{Institute of Veterinary Pathology, Freie Universit{\"a}t Berlin, Berlin, Germany}
\affil[3]{Department of Infectious Diseases and Pathobiology, Vetsuisse Faculty, University of Bern, Bern, Switzerland}
\affil[4]{Vet Med Labor GmbH - Division of IDEXX Laboratories, Ludwigsburg, Germany}

\affil[*]{marc.aubreville@fau.de}

\newcommand{\etal}{\textit{et al.~}}

\begin{abstract}

Manual count of mitotic figures, which is determined in the tumor region with the highest mitotic activity, is a key parameter of most tumor grading schemes. It can be, however, strongly dependent on the area selection due to uneven mitotic figure distribution in the tumor section.

We aimed to assess the question, how significantly the area selection could impact the mitotic count, which has a known high inter-rater disagreement. On a data set of 32 whole slide images of H\&E-stained canine cutaneous mast cell tumor, fully annotated for mitotic figures, we asked eight veterinary pathologists (five board-certified, three in training) to select a field of interest for the mitotic count. To assess the potential difference on the mitotic count, we compared the mitotic count of the selected regions to the overall distribution on the slide.

Additionally, we evaluated three deep learning-based methods for the assessment of highest mitotic density: In one approach, the model would directly try to predict the mitotic count for the presented image patches as a regression task. The second method aims at deriving a segmentation mask for mitotic figures, which is then used to obtain a mitotic density. Finally, we evaluated a two-stage object-detection pipeline based on state-of-the-art architectures to identify individual mitotic figures.

We found that the predictions by all models were, on average, better than those of the experts. The two-stage object detector performed best and outperformed most of the human pathologists on the majority of tumor cases. The correlation between the predicted and the ground truth mitotic count was also best for this approach (0.963 to 0.979). Further, we found considerable differences in position selection between pathologists, which could partially explain the high variance that has been reported for the manual mitotic count. To achieve better inter-rater agreement, we propose to use a computer-based area selection for support of the pathologist in the manual mitotic count.

\end{abstract}
\begin{document}

\flushbottom
\maketitle
\thispagestyle{empty}

\section{Introduction}

Patients with tumors profit significantly from a targeted treatment, and a key to this is the assessment of prognostic relevant factors \cite{Veta:2015bi}. Cells undergoing cell division (mitotic figures) are an important histological parameter for this: It is widely accepted that the density of cells in mitosis state, the so-called mitotic activity, strongly correlates with cell proliferation, which is amongst the most powerful predictors for biological tumor behavior \cite{Baak:2008cm}. Consequentially, mitotic activity is a key parameter in the majority of tumor grading systems and provides meaningful information for treatment considerations in clinical practice \cite{Elston:1991dl,sledge2016canine}. 

For example, the scheme by Elston and Ellis, which is commonly used to assess human breast cancer, proposes the count of mitotic figures within ten standardized areas at $400\times$ magnification (high power field, HPF), resulting in the mitotic count. Prognosis is determined by the mitotic count being between $0-9$, $10-19$, and $>20$ for a low, moderate, and high score with respect to malignancy of the tumor \cite{Elston:1991dl}. The grading system by Kiupel \textit{et al.} for the assessment  of canine cutaneous mast cell tumors (CCMCT), a highly relevant hematopoietic tumor in dogs, requires at least seven mitotic figures per 10 HPF for the classification as high grade, i.e., more malignant, tumor \cite{Kiupel:2011du}.

Common to most grading schemes is the recommendation to count mitotic figures in the area with the highest mitotic density ('hot spot'), which is suspected to be in a highly cellular area in the periphery of the tumor section \cite{Azzola:2003ey,Meuten:2016fa,Veta:2015bi}.
As has been long assumed by many experts, we have recently confirmed for the case of CCMCT that mitotic figures can have a patchy distribution throughout the tumor section \cite{Bertram:2019vp}. Tumor heterogeneity of the mitotic count \cite{Jannink1996,tsuda2000evaluation} and another index for tumor cell proliferation (Ki67 immunohistochemistry) \cite{Focke2016} have also been proven for human breast cancer. 
Opposed to previous assumptions, proliferative hot spots are often not located in the periphery of the tumor section \cite{Bertram:2019vp,staalhammar2016digital}. Due to significant tumor heterogeneity and lack of reproducible area selection protocols, the selection of the area will be influenced by a subjective component and is additionally restricted by limited time in a diagnostic setting. Accurate spotting of mitotic figures requires high magnifications \cite{Meuten:2016fa}, which makes the selection of a relatively small, but most relevant field of interest from a large tumor section difficult for pathologists.

The count of mitotic figures is known to have low reproducibility \cite{Meyer:2005cl,Meyer:2009eu}. While a low inter-rater agreement of mitotic figures will be one reason for some variance in mitotic count between experts \cite{Meyer:2005cl}, the area selected for counting has certainly also a significant influence\cite{Fauzi:2015iw,tsuda2000evaluation}. 
The sparse distribution of mitoses is calling for an increase in the number of HPF to be counted within \cite{Meyer:2009eu,Bonert:2017go}.  Manual count of mitotic figures is, however, a tedious and labor-intensive process, which puts a natural restriction on this number in a clinical diagnostic setting. Additionally, with an increasing number of HPF, the mitotic count (MC) will converge towards the average MC of the slide, which contradicts the idea of assessment in the most malignant area of the tumor, where the result is likely to have the greatest prognostic value. Although a proof of the higher prognostic value of tumor 'hot spots' compared to average counts have not been shown for the MC previously, recent studies have shown that hot spots of the proliferative immunohistochemical marker Ki67 is superior to enumerating entire tumor section or the tumor periphery \cite{staalhammar2016digital,staalhammar2018digital}.  

While the time a pathologist can spend on region selection in a clinical setting is limited and thus only small parts of the tumor can be examined, algorithmic approaches can screen the complete whole slide images (WSI) for mitotic figures within a short time and independent from human interaction. To improve reproducibility and accuracy of the manual mitotic figure count, we thus propose to incorporate an automatic or semi-automatic preselection of the region of the highest mitotic count. This augmentation has the potential to improve prognostic value and reduce time effort for human prognostication at the same time.

In this work, we compared three convolutional neural network (CNN)-based approaches, and assessed their ability to estimate the area of highest mitotic activity on the slide.  The main technical novelty of this work is the customization of three very different state-of-the-art deep learning methods, representing the tasks of segmentation, regression and detection, and their embedding into a pipeline tailoring for this task. On a data set of completely annotated WSI from CCMCT, we compared the algorithms performance against those of five board-certified veterinary pathologists and three veterinary pathologists in training. To the best of the authors' knowledge, this is the first time that human expert were compared in such a task.

\section{Related work}
Mitotic figure detection is a known task in computer vision for more than three decades \cite{Kaman:1984em}. It took until the advent of deep learning technologies \cite{maier2019gentle}, first used by Cire\c{s}an \textit{et al.} \cite{Ciresan:2013upa}, for acceptable results to be achieved. The models used in this approach were commonly pixel classifiers trained with images where a mitotic figure is either at the center (positive sample) or not (negative sample). In recent years, significant advances were made in this field, also fostered by several competitions held on this topic \cite{Roux:2013kn,Veta:2015bi,veta2018predicting}. Especially the introduction of deeper residual networks \cite{He:2016ib} had a tremendous influence on current methods. One example of this is the DeepMitosis framework by Li \textit{et al.} \cite{Li:2018ce}. Their work combines a mitotic figure detection approach based on state-of-the-art object detection pipelines with a second-stage classification and shows that this second stage can significantly increase the performance for the task of mitotic figure classification.

Even though results, as achieved by Li \textit{et al.} on the 2012 ICPR MITOS data set\cite{Roux:2013kn} with F1-scores of up to $0.831$, are impressive, the approaches will likely still not meet clinical requirements for a several reasons: Firstly, robustness to image variability increasing factors like changes in stain or image sharpness were not included in the data set, while this is common in whole slide images (WSI). This is especially true for the 2012 ICPR MITOS data set, where training and test images were extracted from the same WSI \cite{Roux:2013kn}, and thus the test set does not fulfill the purpose of assessment of robustness. Further, only a small portion of the tumor was annotated in this data set (and also all other publicly available data sets). We can thus question the generalization of these results, since the data sets overestimate the prevalence of mitotic figures and specifically exclude structures that have a similar visual appearance to mitotic figures like cellular structures in necrotic tissue or artifacts from the excision boundary of the tumor. 

Pati \etal \cite{Pati:2019kk} have performed algorithmic assessments on the WSI of the TUPAC16\cite{veta2018predicting} data set, which does, however, also not include mitotic figure annotations for locations outside the 10 high power field (HPF) region that was previously selected by a pathologist \cite{Veta:2015bi}. To circumvent these limitations, Pati \etal employ an invasive tumor region mask generation to exclude tissue outside the main tumor in a semi-supervised approach. They introduce a mitotic density map as visualization and add a weighting factor that is inversely proportional to the distance to the tumor boundary \cite{Pati:2019kk}. They were, however, due to the lack of annotation data, not able to verify these maps against some kind of ground truth. The weight given to the mitotic density map strongly relies on the assumption that the tumor border contains more actively metastatic cells than internal regions\cite{Pati:2019kk}. While this assumption is part of many MC protocols, there is no work known to the authors that provides a general proof for this, especially for a greater variance of tumor types. In fact, for the case of CCMCT, a previous analysis \cite{Bertram:2019vp} showed no such relationship on the data set used in this work \cite{scidata}. Also Ki67 hot spots had been found in the periphery in only some cases of human breast cancer \cite{staalhammar2016digital}.

\section{Material}

Our research group built a data set consisting of 32 CCMCT cases, where all mitotic figures have been annotated within the entire tumor area. The data set, including a detailed description of its creation and anonymized forms of all WSI and all annotations, is available publicly \cite{scidata}. All tissue samples were taken retrospectively from the diagnostic archive of the authors with approval by the local governmental authorities (State Office of Health and Social Affairs of Berlin, Germany, approval ID: StN 011/20), and were originally provided from routine diagnostic service for purely diagnostic and therapeutic reasons, i.e. no animal was harmed for the construction of the data set or this study. All methods were performed in accordance with the relevant guidelines and regulations. Tissue section were routinely prepared and slides were stained with standard hematoxylin and eosin (H\&E) dyes using a tissue stainer (ST5010 Autostainer XL, Leica, Germany), prior to being digitized using a linear scanner (Aperio ScanScope CS2, Leica Biosystems, Germany) at a resolution of $0.25\frac{\mu m}{px}$. 

\begin{figure}
\centering
\includegraphics[width=\textwidth]{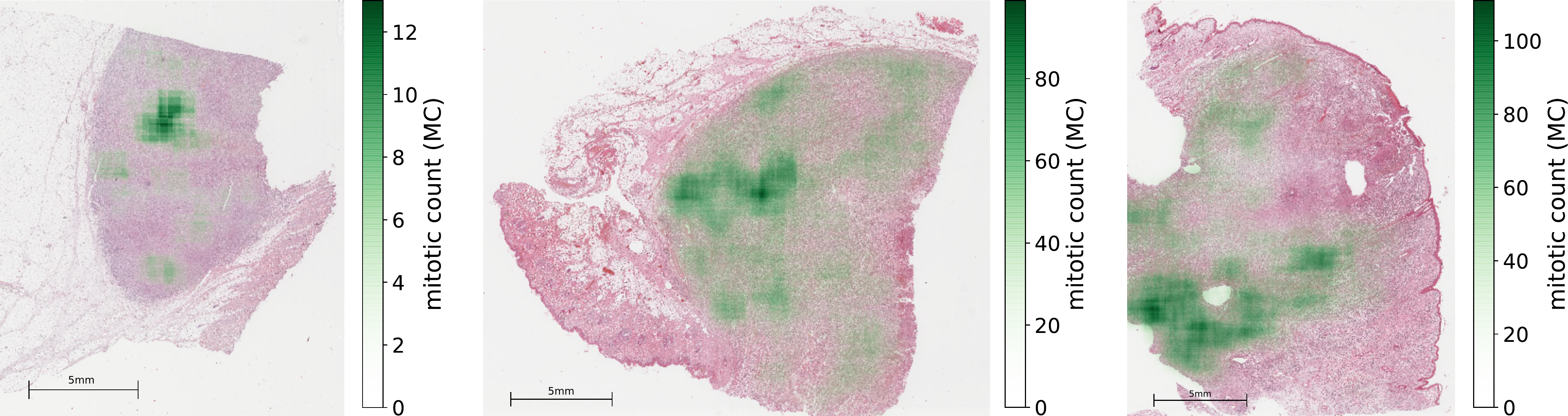}
\caption{Distribution of mitotic count (count of mitotic figures per 10 HPF area), represented as a green overlay over the H\&E-stained tumors from the CCMCT data set.}
\label{mitdensity}
\end{figure}

Using a novel software solution \cite{sliderunner}, it was possible to build up a database that includes annotations for both true mitotic figures and look-alikes that might be hard to differentiate. 
The initial screening was performed twice by the first pathologist, who annotated unequivocal mitoses, possible mitosis, and similar structures. The second pathologist then provided an additional label for all annotated cells. For this, the second export did not know about the class of the cells, nor about the distribution of classes. This setup was chosen in order to limit the bias in the data set acquisition while at the same time providing a very high sensitivity for selecting the potential mitotic figures.

In a follow-up step, both experts were presented with disagreed cells and found a common consensus. In order to incorporate potentially missed mitotic figures, we employed an deep-learning based object-detection pipeline to find additional mitotic figure candidates, not being part of the data set before, which were subsequently also assessed by two pathologists with a final joint consensus. In this step, we used a low cutoff for the machine learning system to enable high sensitivity, leading to 89,597 candidates, which finally resulted in only 2,273 mitotic figures confirmed by both pathologists. This procedure enabled the generation of a high-quality mitotic figure data set that is unprecedented in size to date, including a total of 44,880 mitotic figures and 27,965 non-mitotic cells with similar appearance to mitotic cells (hard negatives). The total tumor area in all 32 cases is $4,939\,mm^2$ ($\mu=149.68\,mm^2, \sigma=96.99\,mm^2$). 

This novel data set provides us, for the first time, with the possibility to assess the performance of algorithms for a region of interest prediction at a large scale. Using this data set, we can derive a (position-dependent) ground truth mitotic count (MC) by counting all mitotic figures in a window of 10 HPF size around the given position, and thus evaluate how the MC depends on the position used for counting. To perform cross-validation, we split up the slide set into three batches, where two would be used for training and validation (model selection), and one would serve as a test set.

An analysis of this data set underlines the assumption that the distribution of mitotic figures is not uniform but rather patchy (see Figure~\ref{mitdensity}), which further highlights the need for a proper selection of the HPFs for a reproducible manual count.

\section{Methods}
In this work, our aim was to compare field-of-interest (FOI) detection for algorithmic approaches versus the manual selection by pathologists in CCMCT histological sections. 
We were unable to compare this on other publicly available data sets 
since no ground truth mitotic figure annotation data is available for the entire whole slide image.

\subsection{Pathology expert performance evaluation}
To set a baseline for the task, we asked five board-certified veterinary pathologists (BCVP) and three veterinary pathologists in training (VPIT) to mark the region of interest (spanning 10 HPF) which they would select for enumerating mitotic figures. The experts came from three different institutions. For this, we set no time limit, but pathologists were instructed to act as they would for routine diagnostics. Since the area of a single HPF is varying according to the optical properties of the microscope, we chose $2.37\,mm^2$ as the standard 10 HPF area as recommended by Meuten \cite{Meuten:2016fa} at an aspect ratio of 4:3. While this area might not be sufficient for a stable classification, as shown by Bonert and Tate \cite{Bonert:2017go}, it is currently the recommended area size in veterinary pathology. From the $2.37mm^2$ area selected by the pathologist, we calculated the \textit{ground truth} mitotic count, which was available with a two-expert-consensus from annotations in the data set. 

\begin{figure*}
	\includegraphics[width=\linewidth]{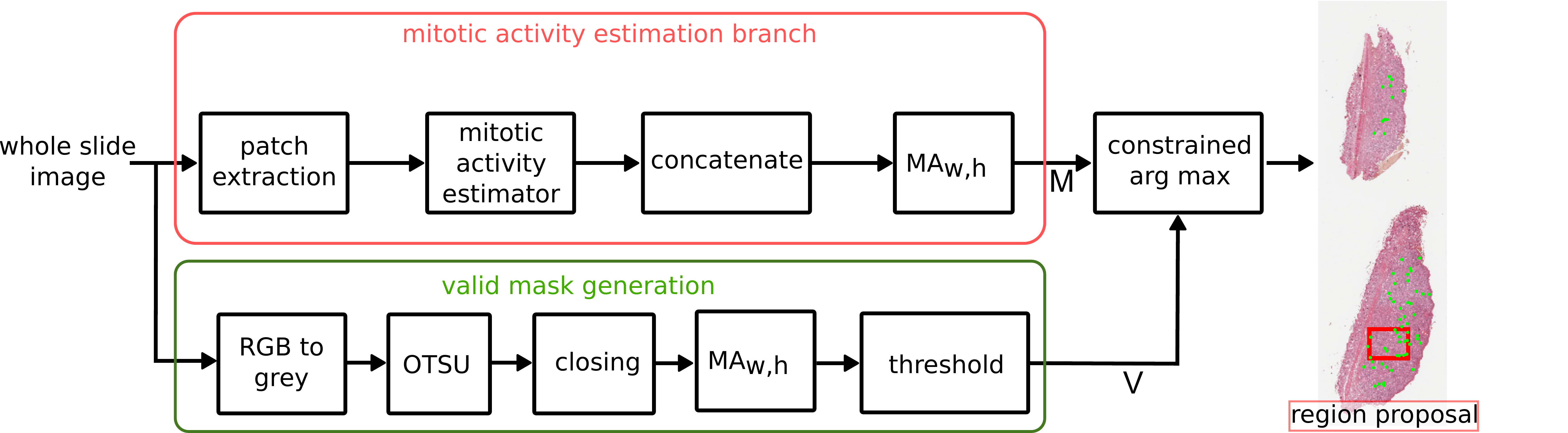}
	\caption{Overview of the general framework. The CNN-based mitotic density estimator is applied to each WSI image patch. After the calculation, a moving window averaging operation yields mitotic density over the area of 10 high power fields. Adapted from \cite{Aubreville:2019um}.}	
	\label{overview}
\end{figure*}

\subsection{General algorithmic approach}
We compared three state-of-the-art methods, all aiming at the prediction of the mitotic count within a defined area of a histology slide. The first followed the indirect approach of predicting a mitotic figure segmentation map, while the second tried to estimate the number of mitotic figures within an image directly, and the third tried to detect mitotic figures as objects using a pipeline inspired by the works of Li \etal \cite{Li:2018ce}. For all methods, the original WSI was split up into a multitude of single images that were subsequently processed using the CNN. The result was concatenated to yield a scaled estimate for the mitotic density (see Figure~\ref{overview}). All approaches were embedded into a toolchain where the estimator is followed by a concatenation and a 2D moving average (MA) operation. The filter kernel of this operation is in accordance with the width and height of the FOI (10 HPF). After the MA operator, the position of the maximum value within a valid mask $V$ is determined as the center of the region of interest.
The valid mask generation pipeline (lower branch of Figure~\ref{overview}) performs a thresholding operation, followed by morphological closing. This gives a reasonable estimate of slide area filled by tissue.
To exclude FOI predictions in border areas of the slide that are not to at least  $95\,\%$ covered by tissue, we perform an MA operator of the same dimensionality followed by a threshold of $0.95$, yielding the valid mask $V$ (see Figure~\ref{overview}).

\subsection{Estimation of mitotic figures via object detection}
We took inspiration from the approaches by Li \etal \cite{Li:2018ce} and built a dual-stage object detection network for this task (see Figure~\ref{unetfig}~a). Over integrated dual-stage algorithms like Faster R-CNN \cite{Ren:2017kt}, this is only a minor additional computational effort (since the second stage is only processed for a detections by the first stage). The first stage is a customized RetinaNet-architecture \cite{marzahl2019deep,Lin:2017de}, incorporating a ResNet18 \cite{He:2016ib} stem with feature pyramid network (FPN), of which we only used the layer with the highest resolution, and two customized heads (one for classification and one for position regression). The second stage is taking extracted cropped patches centered around a potential mitotic figure coordinate as an input and uses a standard ResNet18-based CNN classifier to perform a refined classification. 
As also described by Li \etal, we found the second stage to increase performance significantly\cite{scidata}.

\subsection{Estimation of mitotic count using segmentation}

\begin{figure*}[hbt]
	\includegraphics[width=\textwidth]{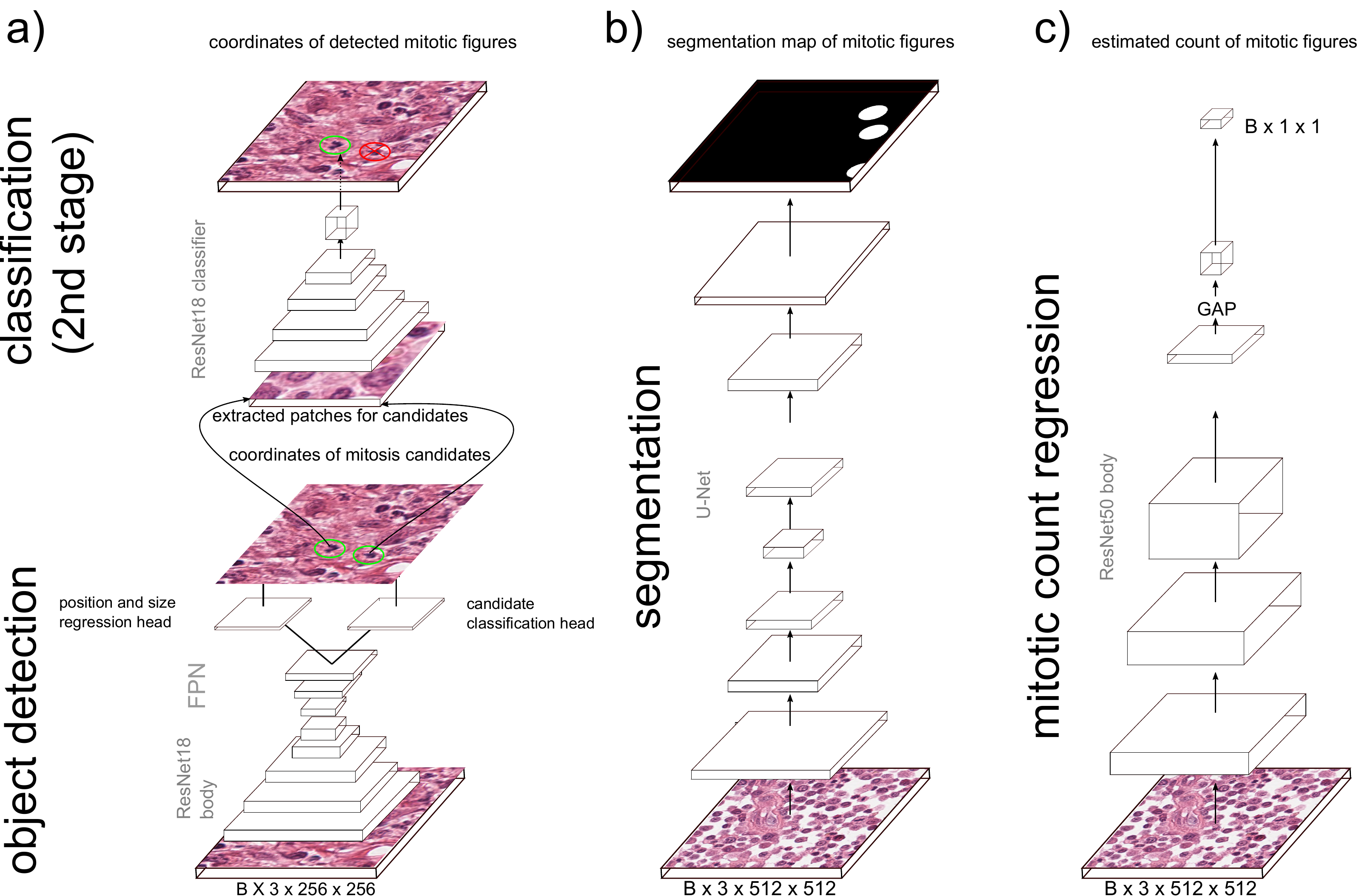}
	\caption{Different mitotic figure density estimation approaches investigated in this work. a) Dual-stage object detection pipeline based on a custom configuration of RetinaNet \cite{Lin:2017de} and a ResNet-18 \cite{He:2016ib}-based classifier. b) Generation of a mitotic figure segmentation map using a U-Net \cite{Ronneberger:2015gk} network architecture, as described in \cite{Aubreville:2019um}. c) Estimation of mitotic density using a direct regression approach based on a ResNet-50 architecture.}
	\label{unetfig}
\end{figure*}

While mitotic figure detection is usually considered an object detection task, where the position and the class probability of an object are estimated, we wanted to estimate a map of mitosis likelihoods for a given image. One approach that has been successfully used in a significant number of segmentation tasks is the U-Net by Ronneberger \textit{et al.} \cite{Ronneberger:2015gk}. As previously shown, this approach can be utilized well for the given task \cite{Aubreville:2019um}. The target map for the network consists of filled circles of 50px diameter, wherever a mitotic figure was present (see Figure~\ref{unetfig}~b). We assume that this shape is a good estimate to represent the wide range of possible appearances a mitotic figure can have. 

\subsection{Regression of the mitotic figure count within an image patch}
\label{WSL}

Direct estimation of the mitotic figure count is the most straight-forward way of deriving the area with the highest mitotic activity in the slide. We use a convolutional neural network with a single output value to regress the (normalized) mitotic count for each inference run. We based the network architecture of this approach on a ResNet50 stem \cite{He:2016ib}, to which we attached a new head consisting of a double layer of $3\times3\times128$ convolutions and batch normalization, followed by a global average pooling layer (GAP) and a final $1\times1\times1$ convolutional layer with sigmoid activation (see Figure~\ref{unetfig}~c).

To define the count of mitotic figures in an image, including partial mitotic figures in the border regions of the image (with width $w$ and height $h$), we denote the x and y position of each mitotic figure in coordinates relative to the image center as $p_x$ and $p_y$, respectively. The approximated diameter of a mitotic figure is denoted $d$. The overall count, i.e., the target for the regression, is then defined as:

\begin{equation}
	C = \frac{1}{\beta}\sum_i{\frac{\gamma(\left|p_x(i)\right|-\frac{w}{2})\gamma(\left|p_y(i)\right|-\frac{h}{2}) }{d^2}}
\end{equation} 

where the partial weight $\gamma(x)$ in dependency of the positional offset is defined as:
\begin{equation}
\gamma(x) = \left\{ 
	\begin{array}{l l}
			d & \quad x < \frac{d}{2} \\
			\frac{d}{2}-x & \quad 0 \leq x < d \\
			0 & \quad x > \frac{d}{2}
	\end{array}
	\right.
\end{equation}

The normalization parameter $\beta$ was chosen heuristically as $\beta=10$. As loss function, we use the quadratic error ($L^2$) for each image.

\subsection{Training and model selection}
Besides the train-test split mentioned, we divided the training WSI into a training portion (upper 80\% of the WSI) and a validation portion (remainder). For training, we used a sampling scheme that would draw images randomly from the slide, according to three group definitions:

\begin{enumerate}
	\item Images containing at least one mitotic figure.
	\item Images containing at least one non-mitotic figure (hard negative example), as annotated by pathology experts.
	\item Images randomly picked on a slide without any further conditions.
\end{enumerate}

Each group of images had a dedicated justification of inclusion: While we included images with at least one mitotic figure for network convergence, the inclusion of hard negatives is well known in literature, although commonly in the form of hard example mining from the data set \cite{veta2018predicting}, which we, however, don't need to employ since the data set provides these for our purpose. Finally, we include random picks to also train the network on the tissue where no mitotic figure can be expected. This is especially interesting for necrotic tissue, where nuclei can have a mitosis-like appearance and for cells near the excision boundary of the tumor.
All images were randomly sampled from the training WSI using the criteria described above and were additionally rotated arbitrarily prior to cropping. Since WSI are typically very large (in the order of several Gigapixels), this resulted in a vast number of combinations. For validation, after 10,000 iterations, a completely randomly drawn batch of 5,000 images was fed to the network. The training batch severely overestimates the a priori probability of a mitotic figure being present, but we found that this setup speeds up model convergence significantly. 
After around 1,500,000 single iterations, the models had typically converged.
Even though the choice of images alleviates the heavily skewed distribution, the loss function still needs to be chosen appropriately to cater to imbalanced sets. For this reason, we used a negative intersection over union as loss \cite{Rahman:2016dk} for the U-Net approach. L2 was used for the regression approach. We trained the model using the ADAM adaptation method in TensorFlow with an initial learning rate of $10^{-3}$. 
For model selection, we used the model state, which yielded the highest F1-score during validation for the U-Net approach. The model with the least mean squared error in prediction was chosen for the regression approach. For the object detection-based approach, we used a modestly differing training scheme, which was already described in a previous publication\cite{scidata}.

\section{Results}

\begin{figure}[t]
    \centering
	\includegraphics[width=0.8\textwidth]{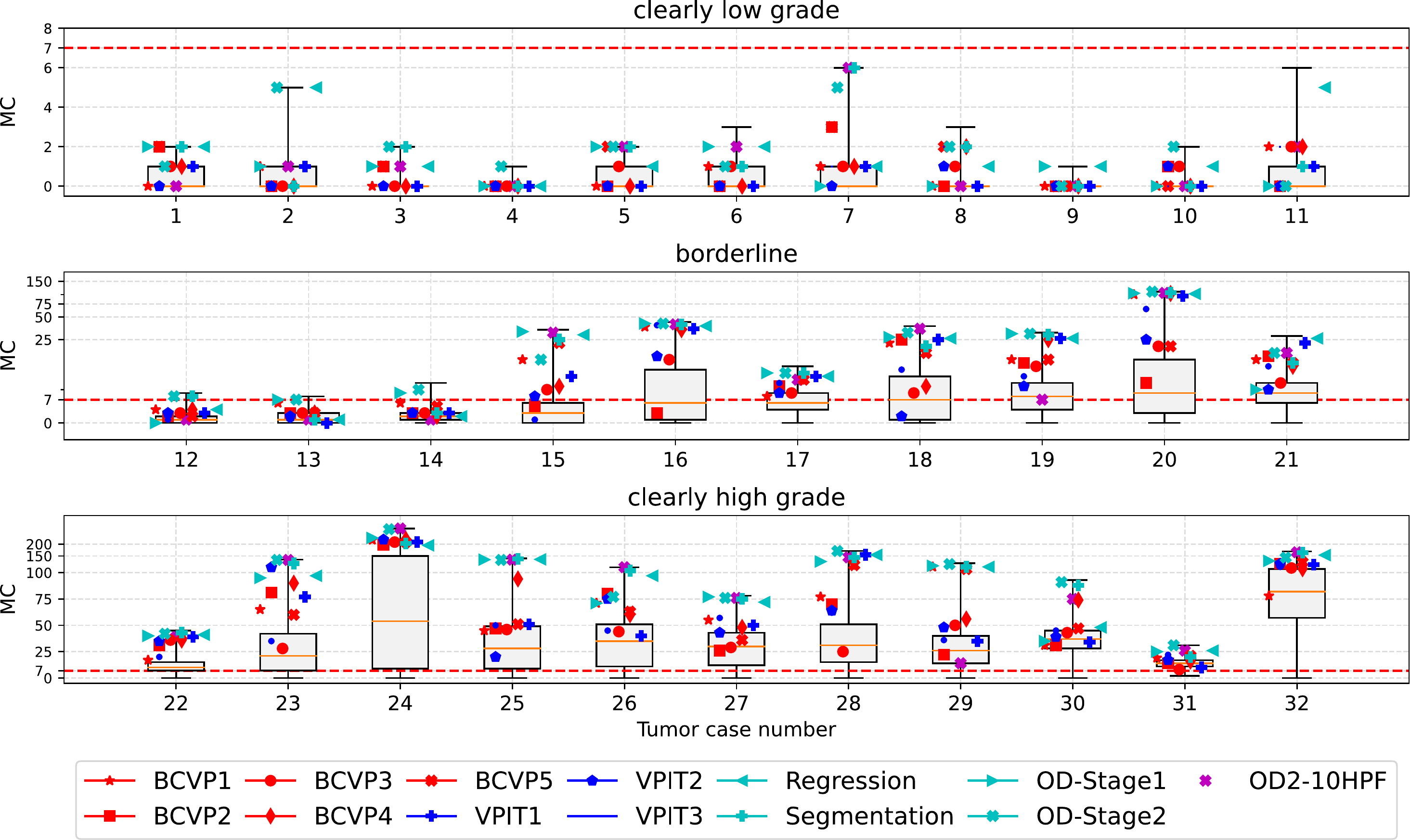}
	\caption{
	Results of field-of-interest selection with tumor cases grouped according to previous work\cite{Bertram:2019vp}. The box-whisker-plots visualize the distribution of mitotic count as per ground truth from the annotations of the data set. Markers indicate mitotic count derived from the positions selected by the pathologists or the algorithms. Since finding the area of the absolute maximum mitotic count is the objective, higher values are better. Results shown are derived from areas selected by five board-certified veterinary pathologists (BCVP), three veterinary pathologists in training (VPIT) and the five algorithmic approaches (regression, segmentation, first stage object detector (OD-Stage1), second stage object detector (OD-Stage2) and ablated data set for second stage object detector (OD2-10HPF)). The red dotted line indicates the threshold as per Kiupel's grading scheme\cite{Kiupel:2011du}.}  
	\label{Results_Boxplot}
\end{figure}

\begin{figure}[hbt]
	\centering
	\includegraphics[height=5.5cm]{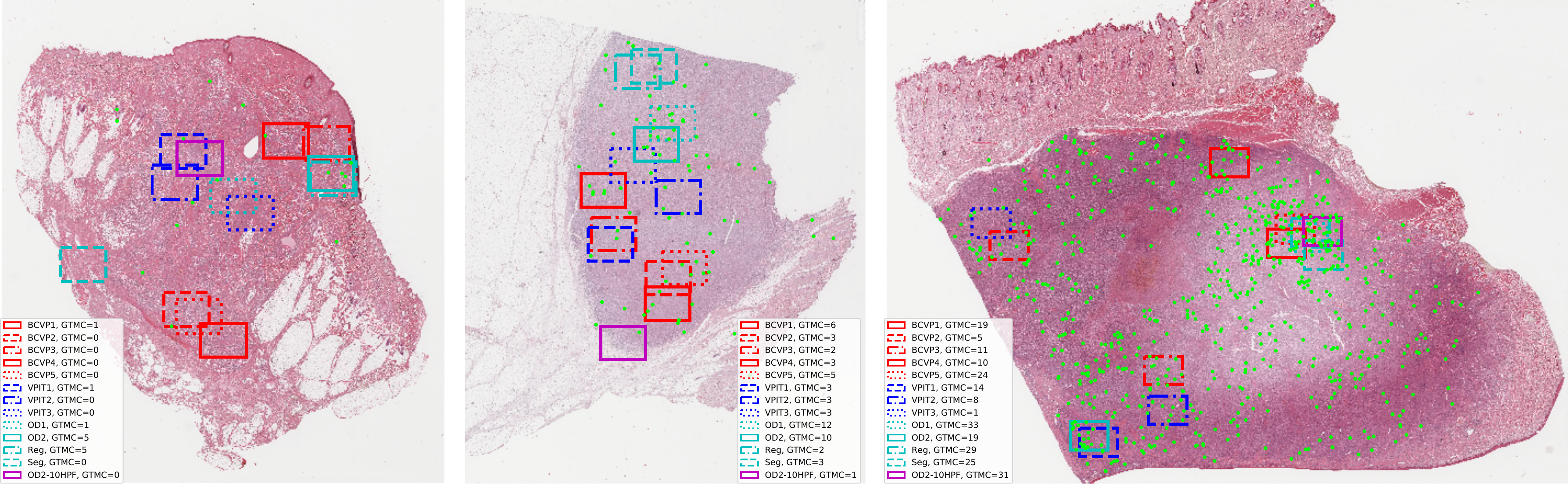}
	\caption{Tumor cases 2 (left), 14 (middle), and 15 (right) with mitotic figures as green points. Markings indicate the selected area of interest. Different choices of the area by the pathologists led to substantial differences in the count of mitotic figures (based on ground truth mitotic count, GTMC) in that area.}
	\label{resultexamples}
\end{figure}

The comparison of the mitotic count (as per ground truth annotation) at the predicted and pathologist-selected positions shows considerable differences. For better visualization, the tumor cases have been grouped into clearly low grade (all possible counts with MC < 7), borderline mitotic count, and clearly high grade (more than 75\% of the possible counts with MC $\geqq$ 7), as in a previous study \cite{Bertram:2019vp}. The spread is most visible in the group of tumors with borderline MC (middle panel of Figure~\ref{Results_Boxplot}). In this group, the maximum of the mitotic count (i.e. the MC at the most mitotically active region) is above 7, which is the threshold in Kiupel's scheme, yet the 25th percentile of the distribution is below this threshold. Thus, by definition, all of these tumors are (at least considering MC as sole criteron) high-grade tumors, yet with a non-negligible risk for inappropriate region selection leading to potential underestimation of the grade \cite{Bertram:2019vp}. Thus, in this group, the area selection has a considerable impact on the mitotic count being above or below the threshold. This is most obvious for the cases 12-16, and 18, where at least one pathologist selected an area for grading that did not have an above-threshold mitotic count. Cases 2, 14, and 15 are also shown in detail in Figure~\ref{resultexamples}, highlighting the effect of the selection of different areas on the mitotic count. In three all cases, the pathologists picked regions of high cellularity, as recommended by the grading scheme \cite{Kiupel:2011du}, yet mitotic figures were not evenly spread in this tissue, leading to a large spread in mitotic count in all three cases. For case 14, the region of the highest mitotic count was not found by any of the pathologists, and regions selected contained at maximum half the number of mitotic figures compared to the maximum area identified by one of the algorithmic approaches. A two-way ANOVA revealed no significant difference between the group of board-certified veterinary pathologists (BCVP) and the group of veterinary pathologists in training (VPIT) (F=0.1358, p(F)=0.7129 for being in the 'board-certified' /'in-training' group, F=71.465, p=0.0000 for the tumor case). Considering the question of FOI-derived mitotic count being below or above threshold ($\geq7$, as per Kiupel's scheme \cite{Kiupel:2011du}), we find a substantial agreement ($\kappa=0.713$) for the VPIT group and an almost perfect agreement ($\kappa=0.822$) for the BCVP group for the group of borderline cases, and a perfect agreement for all other cases. Both test were performed with python statsmodels package, version 0.10.1.

\subsection{Algorithmic estimation of mitotic count}
The number of mitotic figures varies strongly not only with the size of the region of interest but also with its position. We always used an area of 10 HPF ($2.37\,mm^2$), as recommended by Kiupel \etal. For the purpose of selecting the mitotically most active 10 HPF region, it is not of interest if the algorithmic approach over- or underestimates the likelihood of mitotic figures, as long as this over- or underestimation is consistent over the whole data set. As such, it is similar to a scaling factor or an offset. We expect, however, to have a high correlation between the mitotic count estimated and the ground truth mitotic count. For this reason, we chose to use the correlation coefficient as one measure to evaluate the approaches.

In general, we find high correlations of the predicted mitotic count by the algorithmic approaches (see Table \ref{results_table}). Of all the described approaches, clearly, the two-stage object detection pipeline shows the best performance, with correlations between $0.963$ and $0.979$, depending on the cross-validation fold. We do not find big differences in the single-stage approaches (segmentation, mitotic count regression, and single-shot object detection) in this metric. These correlations also translate into generally good performance when trying to find the mitotically most active region on the complete whole slide images (Figure~\ref{Results_Boxplot}). The group of algorithmic approaches is, on average, better than the pathologist group for every single case. Also from this evaluation, we find that the dual-stage object detection pipeline (indicated by a cyan cross) scores best, achieving highest mitotic count on 16 out of 32 tumor cases (2-4, 8, 10, 12, 17, 19, 20, 22-23, 25, 28-31), and being outperformed by a pathologist only in five cases (1, 5, 11, 15, 26).

\begin{table}
\centering
\begin{tabular}{|c|c|c|c|c|c|c|}
	\hline
	Method & \multicolumn{3}{|c|}{Full Data Set}& \multicolumn{3}{|c|}{Ablation Study} \\
Cross-val & fold 1 & fold 2 & fold 3 & fold 1 & fold 2 & fold 3 \\
\hline
Segmentation & 0.925 & 0.947 & 0.963 & - & - & - \\
MC Regression & 0.924 & 0.935 & 0.948 & - & - & - \\
Object Detection, 1 stage & 0.916 & 0.939 & 0.957 & 0.868 & 0.904 & 0.930  \\
Object Detection, 2 stage & 0.963 & 0.971 & 0.979 & 0.897 & 0.924 & 0.944  \\
	\hline
\end{tabular}	
\caption{Correlation coefficient between ground truth and estimated mitotic count on all evaluated models, data sets, and all folds of the cross-validation. We look for the position of the highest mitotic density. Thus correlation is a suitable measure to compare models, as scaling and additional factors have no influence in the subsequent argmax operation.}
\label{results_table}
\end{table}

\subsection{Ablation study on 10HPF data set}

All previously published data sets with annotations \cite{veta2018predicting,Veta:2014gu,roux2014mitos,Roux:2013kn} on mitotic figures only considered an area selected by an expert pathologist, typically spanning 10 HPF. The TUPAC16 data set was the first to provide the complete WSI additionally, yet, without annotations outside the field of interest that was selected by the pathologist. 
Veta \textit{et al.} state that the task of finding such a region of interest for the mitotic count is certainly a task worth automating\cite{Veta:2015bi}. Since our data set provides us with the possibility to assess algorithmic performance on this task, we asked ourselves the question of how much our best performing (two-stage) algorithmic pipeline would degrade if we trained it with a limited data set, also only spanning 10 HPF. We aim to relate to the state-of-the-art in data sets with this methodology.

\begin{figure}[hbt]
    \centering
    \includegraphics[width=1.0\textwidth]{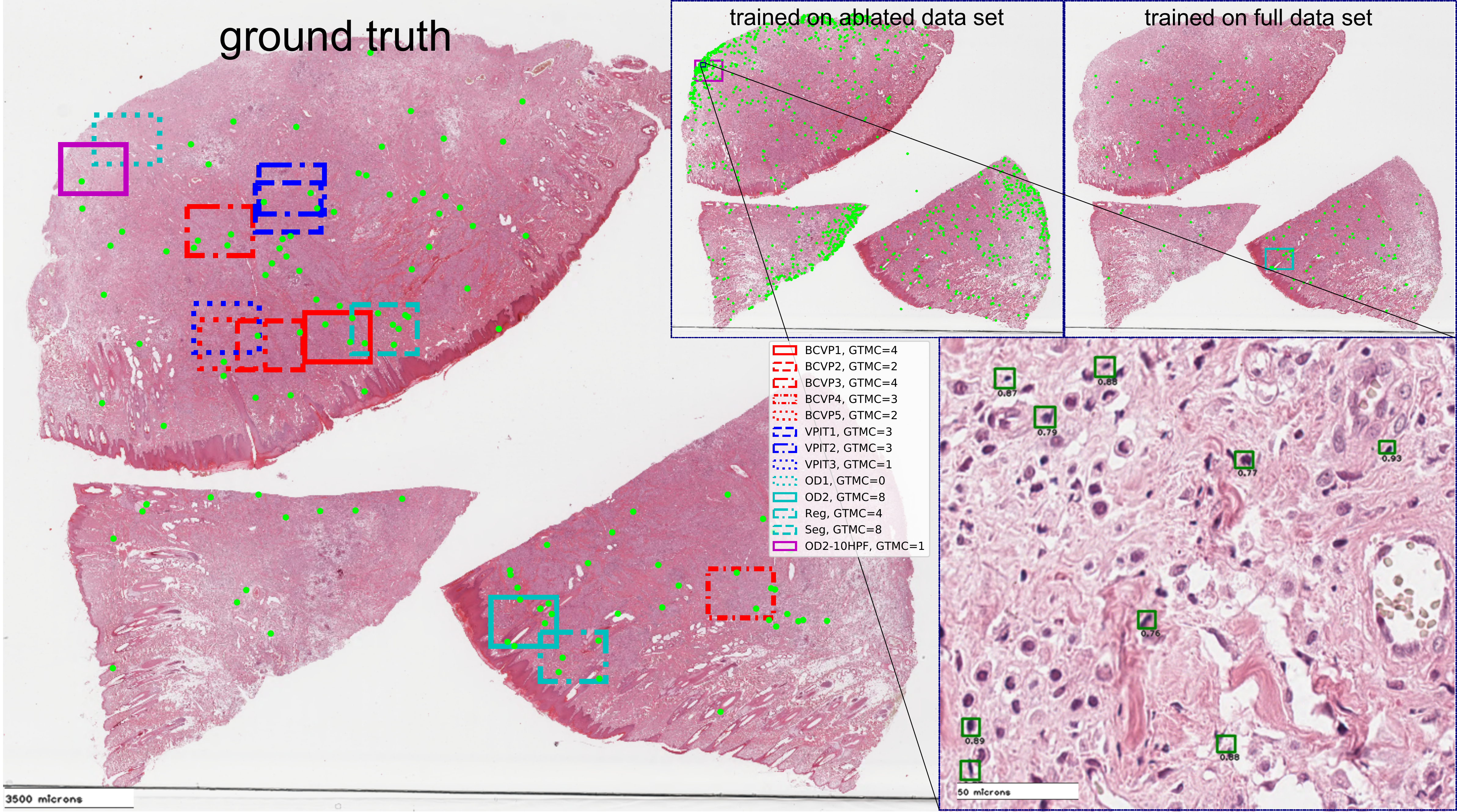}
    \caption{Ablation study on data set size, case 12. Ground truth (left), estimate by dual-stage object detection pipeline trained on reduced (top middle) or full (top right) data set. Annotations and detected mitotic figures by two-stage object detection pipelines are given as green dots in overview and as green rectangles in detail view (bottom right). Expert and model selections are given as rectangles spanning an area of 10 HPF.}
    \label{fig:ablation}
\end{figure}

We find a significantly reduced correlation of the estimated mitotic count with ground truth data in this case (see Table \ref{results_table}), which also reflects in reduced performance, especially for the borderline tumor cases. For the example of case 14, we can see the influence of deteriorated tissue found at the excision boundary of this tumor section. While in the tissue border, there is a rather low number of true positives (cf. Figure~\ref{fig:ablation} left panel), we find a vast number of false positives related to cells with similar appearance than actual mitotic figures (Figure~\ref{fig:ablation} top middle and bottom right panel), which is not present when trained on the full data set (Figure~\ref{fig:ablation} top right). The ablation study thus suggests that other regions (outside the FOI) are included in the training to allow for generalization to the whole WSI.

\begin{figure}[hbt]
    \centering
    \includegraphics[width=1.0\textwidth]{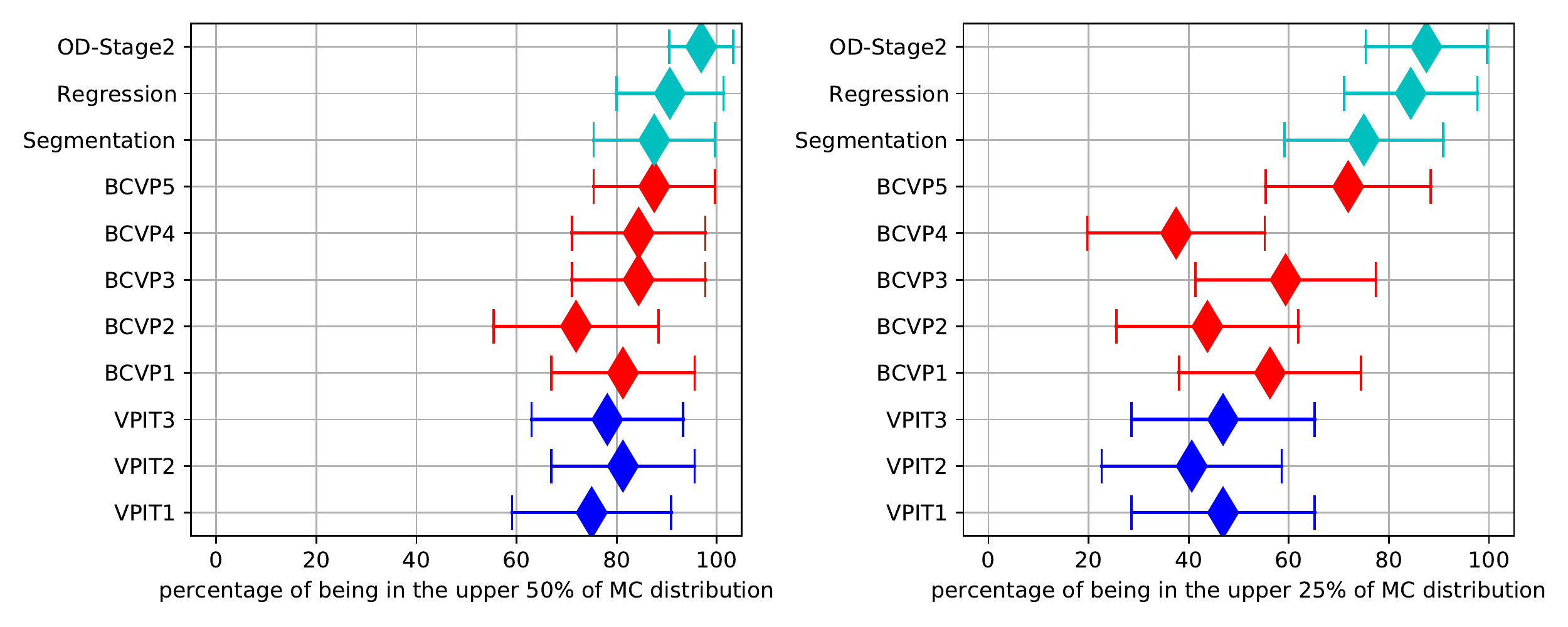}
    \caption{Forest plots of the mitotic count (MC) derived from the region selected 
    being above or equal to the upper half or upper 25\%, respectively, of ground truth MC distribution comparing the main algorithmic approaches (regression, segmentation and object detection), the five board-certified veterinary pathologists (BCVP), and the three  veterinary pathologists in training (VPIT). Markers indicate median values, lines indicate 95\% confidence interval. }
    \label{fig:forestplot}
\end{figure}

\section{Discussion}

Our results support the hypothesis that one significant reason for high rater disagreement in the mitotic count, which is well-known in human and veterinary pathology, lies in the selection of HPFs used for counting. On our data set, we found the distribution of mitoses to be fairly patchy, and selection of area does thus have a substantial impact on the mitotic count. This emphasizes the importance of this preselection task, which was not tackled in any mitosis detection challenge so far. 

Our work also underlines that while pathologists will likely not be consistently able to select the area of highest mitotic activity (see Figure \ref{fig:forestplot}), an algorithmic evaluation of the WSI could prove to be a good augmentation method. 
We have evaluated the tumor selection based on intratumoral heterogeneity within the same tumor section representing a single 2D-plane of the tumor, however, there may also be intratumoral heterogeneity at the 3D-levels, as has been shown for human breast cancer \cite{Jannink1996}. Whereas manual tumor screening is restricted a small number of tumor sections, automated or semi-automated analysis has the potential to significantly increase the number of evaluated tumor sections and thereby allowing a more extensive, systematic (stereological) sampling of the tumor biopsy.  
Besides finding the area of highest mitotic activity, the methods presented in this paper can also well serve as an aid to navigating the whole slide image and could thus generate further insights for a more precise prognosis.

A limitation of current grading ambitions is the poor consistency of the applied grading methods between the different systems. Whereas most grading systems require a mitotic count in ten consecutive, non-overlapping HPF in the area with the highest mitotic density \cite{Azzola:2003ey,Elston:1991dl,sledge2016canine} other systems count mitotic figures in three fields \cite{JKirpensteijn:2009wz} or propose a random selection of individual HPF \cite{JKirpensteijn:2009wz,Loukopoulos:2007dc}. To the authors' best knowledge, current literature does not imply which method has the best agreement with prognosis, and the most informative method and area could also be subject to the tumor type. In the present work, we have therefore decided to use the criteria of the respective grading system, i.e., ten consecutive HPF in the area within the highest density and a size of the HPF in consensus with current guidelines \cite{Meuten:2016fa}. A potential limitation is that we had to preset the aspect ratio of the FOI. We assume that while the actual shape of this area might play a role in an individual case, on average, the impact is negligible. The effect of this, however, will have to be investigated further for data sets where prognostic data is available.
Another potential limitation of the work is that we evaluated deep learning-based approaches on a dataset that was also partially generated with the help of deep learning. The increase of mitotic figures by this augmentation, however, was only $5.3\%$ (42,607 vs. 44,880), and additionally each mitotic figure candidate identified by the system was judged independently by two pathologists. We can thus assume that the potential bias of this augmentation stage on the results of this work is negligible.

Reproducibility is a crucial aspect of each diagnostic method, as also pointed out by Meuten \cite{Meuten:2018ir}. A clear definition of the criteria and methods used for counting, which is essential in this regard, should thus try to reduce individual factors as much as possible. As such, also a manual pseudo-random selection of areas should be questioned due to non-existent reproducibility. While the computer-aided methods may have limitations and not be error-proof, they can increase inter-rater concordance on average if they deliver better performance than the average human expert. 

The non-existence of large-scale data sets of fully annotated whole slide tumor sections in many biomedical domains poses an important problem to fully algorithmic as well as to computer-aided diagnosis in this regard. Our study shows that there is a considerable potential for novel methods once such data sets also exist for other species and tissues. Our work also indicates that current data sets that feature only annotations for a very limited area within the slide might show severe limitations when algorithmic solutions shall be transformed into real applications within the pipeline of the pathologist. As suggested by our ablation study, results on current data sets might not be representative of entire tumor sections, and could even lead into wrong comparisons, as algorithmic increments could not translate into a real-world benefit. Given the importance of this task and the known rater disagreement, this should thus be a strong call for the creation of such data sets also on human tissue.

Current grading schemes were mostly developed in retrospect \cite{Elston:1991dl,sledge2016canine,Meuten:2018ir}, with consideration of the survival rates of the patients and the manual mitotic count. 
Since the slides were not investigated for mitotic figures within the complete tumor area, the mitotically most active region was necessarily unknown and was possibly missed. 
Considering our findings, it is possible that those grading schemes have been based on false low MCs.
Novel methods like those presented in this paper can thus help to develop more precise and reproducible grading schemes, which will then likely have adjusted thresholds.

Even though most grading schemes recommend the count of mitotic figures in the mitotically most active (hot spot) area, there is no general evidence that this leads necessarily to higher prognostic value. Bonert and Tate \cite{Bonert:2017go} and also Meyer \etal \cite{Meyer:2009eu} have called for an increase in area to achieve more meaningful cutoff values, which should improve reproducibility of the mitotic count. Using computer augmentation, an extension of the FOI may in fact become feasible for clinical routine. The \textit{sweet spot} for prognostic value, however, is yet to be evaluated and likely also dependent on at least the tumor type. While taking the mean MC of the complete WSI would improve reproducibility of the MC, it is unclear if the prognostic accuracy would be improved by this apporach.  Stålhammar \etal have recently shown for human breast cancer that hot spot tumor proliferation, based on the Ki-67 index,  provide, in fact, a higher prognostic value than the tumor average or tumor periphery  \cite{staalhammar2018digital}. This study, however, has yet to be confirmed, also for more tumor types. The methods proposed in this work can be utilized for such investigations on mitotic figures in H\&E-stained slides. 

While the advances by deep learning seem remarkable, we are aware that deep learning-based algorithms are neither traceable nor are they necessarily robust. This poses a major problem if applied in the field of medicine in general. To assure not only reliability, but also enable liability, it is important that each intermediate result of the pipeline can be verified by an expert. 

In this regard, especially the object detection-based approach enables the pathologist to verify the detected mitotic figures, and thus observe if the machine learning system worked reliably. A computer+veterinarian solution (computer-assisted prognosis) thus yields the best overall combination of speed, accuracy and robustness, at least for the short- to mid-term future. 

\section*{Author contributions}
M.A. and C.A.B. wrote the main manuscript. M.A. conducted the data experiments. M.A. and C.M. wrote the code of the experiments. C.G., M.D., A.S., F.B., S.M., M.F., O.K.,  and R.K. provided their expert knowledge in the expert assessment experiment. R.K. and A.M. provided essential ideas for the study. All authors reviewed the manuscript.

\section*{Competing interests}
The authors declare no competing interests.

\section*{Compliance with ethical standards}
All procedures performed in studies involving animals were in accordance with the relevant guidelines and regulations of the institution at which the studies were conducted. No IRB approval was required for this study, since all animal tissue was acquired in routine diagnostic service and for purely diagnostic and therapeutic reasons. This work did not involve experiments with humans or human tissue. 

\section*{Acknowledgements}
C.A.B. gratefully acknowledges financial support received from the Dres. Jutta \& Georg Bruns-Stiftung f\"ur innovative Veterin\"armedizin.

\section*{Data availability}
All code and data are available online at \url{https://github.com/maubreville/MitosisRegionOfInterest}.


\begin{thebibliography}{10}
\urlstyle{rm}
\expandafter\ifx\csname url\endcsname\relax
  \def\url#1{\texttt{#1}}\fi
\expandafter\ifx\csname urlprefix\endcsname\relax\def\urlprefix{URL }\fi
\expandafter\ifx\csname doiprefix\endcsname\relax\def\doiprefix{DOI: }\fi
\providecommand{\bibinfo}[2]{#2}
\providecommand{\eprint}[2][]{\url{#2}}

\bibitem{Veta:2015bi}
\bibinfo{author}{Veta, M.} \emph{et~al.}
\newblock \bibinfo{journal}{\bibinfo{title}{{Assessment of algorithms for
  mitosis detection in breast cancer histopathology images}}}.
\newblock {\emph{\JournalTitle{Med Image Anal}}} \textbf{\bibinfo{volume}{20}},
  \bibinfo{pages}{237--248} (\bibinfo{year}{2015}).

\bibitem{Baak:2008cm}
\bibinfo{author}{Baak, J. P.~A.} \emph{et~al.}
\newblock \bibinfo{journal}{\bibinfo{title}{{Proliferation is the strongest
  prognosticator in node-negative breast cancer: significance, error sources,
  alternatives and comparison with molecular prognostic markers}}}.
\newblock {\emph{\JournalTitle{Breast Cancer Res Tr}}}
  \textbf{\bibinfo{volume}{115}}, \bibinfo{pages}{241--254}
  (\bibinfo{year}{2008}).

\bibitem{Elston:1991dl}
\bibinfo{author}{Elston, C.~W.} \& \bibinfo{author}{Ellis, I.~O.}
\newblock \bibinfo{journal}{\bibinfo{title}{{pathological prognostic factors in
  breast cancer. I. The value of histological grade in breast cancer:
  experience from a large study with long-term follow-up}}}.
\newblock {\emph{\JournalTitle{Histopathology}}} \textbf{\bibinfo{volume}{19}},
  \bibinfo{pages}{403--410} (\bibinfo{year}{1991}).

\bibitem{sledge2016canine}
\bibinfo{author}{Sledge, D.~G.}, \bibinfo{author}{Webster, J.} \&
  \bibinfo{author}{Kiupel, M.}
\newblock \bibinfo{journal}{\bibinfo{title}{Canine cutaneous mast cell tumors:
  A combined clinical and pathologic approach to diagnosis, prognosis, and
  treatment selection}}.
\newblock {\emph{\JournalTitle{Vet J}}} \textbf{\bibinfo{volume}{215}},
  \bibinfo{pages}{43--54} (\bibinfo{year}{2016}).

\bibitem{Kiupel:2011du}
\bibinfo{author}{Kiupel, M.} \emph{et~al.}
\newblock \bibinfo{journal}{\bibinfo{title}{{Proposal of a 2-Tier Histologic
  Grading System for Canine Cutaneous Mast Cell Tumors to More Accurately
  Predict Biological Behavior}}}.
\newblock {\emph{\JournalTitle{Vet Pathol}}} \textbf{\bibinfo{volume}{48}},
  \bibinfo{pages}{147--155} (\bibinfo{year}{2011}).

\bibitem{Azzola:2003ey}
\bibinfo{author}{Azzola, M.~F.} \emph{et~al.}
\newblock \bibinfo{journal}{\bibinfo{title}{{Tumor mitotic rate is a more
  powerful prognostic indicator than ulceration in patients with primary
  cutaneous melanoma}}}.
\newblock {\emph{\JournalTitle{Cancer}}} \textbf{\bibinfo{volume}{97}},
  \bibinfo{pages}{1488--1498} (\bibinfo{year}{2003}).

\bibitem{Meuten:2016fa}
\bibinfo{author}{Meuten, D.~J.}
\newblock \bibinfo{title}{{Appendix: Diagnostic Schemes and Algorithms}}.
\newblock In \emph{\bibinfo{booktitle}{Tumors in Domestic Animals}},
  \bibinfo{pages}{942--978} (\bibinfo{publisher}{John Wiley {\&} Sons, Inc.},
  \bibinfo{year}{2016}).

\bibitem{Bertram:2019vp}
\bibinfo{author}{Bertram, C.~A.} \emph{et~al.}
\newblock \bibinfo{journal}{\bibinfo{title}{Computerized calculation of mitotic
  distribution in canine cutaneous mast cell tumor sections: Mitotic count is
  area dependent}}.
\newblock {\emph{\JournalTitle{Veterinary Pathology}}}
  \textbf{\bibinfo{volume}{57}}, \bibinfo{pages}{214--226}
  (\bibinfo{year}{2020}).

\bibitem{Jannink1996}
\bibinfo{author}{Jannink, I.}, \bibinfo{author}{Risberg, B.},
  \bibinfo{author}{{Van Diest}, P.~J.} \& \bibinfo{author}{Baak, J.~P.}
\newblock \bibinfo{journal}{\bibinfo{title}{{Heterogeneity of mitotic activity
  in breast cancer}}}.
\newblock {\emph{\JournalTitle{Histopathology}}} \textbf{\bibinfo{volume}{29}},
  \bibinfo{pages}{421--428} (\bibinfo{year}{1996}).

\bibitem{tsuda2000evaluation}
\bibinfo{author}{Tsuda, H.} \emph{et~al.}
\newblock \bibinfo{journal}{\bibinfo{title}{{Evaluation of the Interobserver
  Agreement in the Number of Mitotic Figures Breast Carcinoma as Simulation of
  Quality Monitoring in the Japan National Surgical Adjuvant Study of Breast
  Cancer (NSAS-BC) Protocol}}}.
\newblock {\emph{\JournalTitle{Japanese journal of cancer research}}}
  \textbf{\bibinfo{volume}{91}}, \bibinfo{pages}{451--457}
  (\bibinfo{year}{2000}).

\bibitem{Focke2016}
\bibinfo{author}{Focke, C.~M.}, \bibinfo{author}{Decker, T.} \&
  \bibinfo{author}{van Diest, P.~J.}
\newblock \bibinfo{journal}{\bibinfo{title}{{Intratumoral heterogeneity of Ki67
  expression in early breast cancers exceeds variability between individual
  tumours}}}.
\newblock {\emph{\JournalTitle{Histopathology}}} \textbf{\bibinfo{volume}{69}},
  \bibinfo{pages}{849--861} (\bibinfo{year}{2016}).

\bibitem{staalhammar2016digital}
\bibinfo{author}{St{\aa}lhammar, G.} \emph{et~al.}
\newblock \bibinfo{journal}{\bibinfo{title}{Digital image analysis outperforms
  manual biomarker assessment in breast cancer}}.
\newblock {\emph{\JournalTitle{Modern Pathology}}}
  \textbf{\bibinfo{volume}{29}}, \bibinfo{pages}{318--329}
  (\bibinfo{year}{2016}).

\bibitem{Meyer:2005cl}
\bibinfo{author}{Meyer, J.~S.} \emph{et~al.}
\newblock \bibinfo{journal}{\bibinfo{title}{{Breast carcinoma malignancy
  grading by Bloom-Richardson system vs proliferation index: Reproducibility of
  grade and advantages of proliferation index}}}.
\newblock {\emph{\JournalTitle{Modern Pathol}}} \textbf{\bibinfo{volume}{18}},
  \bibinfo{pages}{1067--1078} (\bibinfo{year}{2005}).

\bibitem{Meyer:2009eu}
\bibinfo{author}{Meyer, J.~S.}, \bibinfo{author}{Cosatto, E.} \&
  \bibinfo{author}{Graf, H.~P.}
\newblock \bibinfo{journal}{\bibinfo{title}{{Mitotic index of invasive breast
  carcinoma. Achieving clinically meaningful precision and evaluating tertial
  cutoffs.}}}
\newblock {\emph{\JournalTitle{Arch Pathol Lab Med}}}
  \textbf{\bibinfo{volume}{133}}, \bibinfo{pages}{1826--1833}
  (\bibinfo{year}{2009}).

\bibitem{Fauzi:2015iw}
\bibinfo{author}{Fauzi, M. F.~A.} \emph{et~al.}
\newblock \bibinfo{journal}{\bibinfo{title}{{Classification of follicular
  lymphoma: the effect of computer aid on pathologists grading}}}.
\newblock {\emph{\JournalTitle{BMC Med Inform Decis}}}
  \textbf{\bibinfo{volume}{15}}, \bibinfo{pages}{115} (\bibinfo{year}{2015}).

\bibitem{Bonert:2017go}
\bibinfo{author}{Bonert, M.} \& \bibinfo{author}{Tate, A.~J.}
\newblock \bibinfo{journal}{\bibinfo{title}{{Mitotic counts in breast cancer
  should be standardized with a uniform sample area}}}.
\newblock {\emph{\JournalTitle{BioMed Eng Onl}}} \bibinfo{pages}{1--8}
  (\bibinfo{year}{2017}).

\bibitem{staalhammar2018digital}
\bibinfo{author}{St{\aa}lhammar, G.} \emph{et~al.}
\newblock \bibinfo{journal}{\bibinfo{title}{Digital image analysis of ki67 in
  hot spots is superior to both manual ki67 and mitotic counts in breast
  cancer}}.
\newblock {\emph{\JournalTitle{Histopathology}}} \textbf{\bibinfo{volume}{72}},
  \bibinfo{pages}{974--989} (\bibinfo{year}{2018}).

\bibitem{Kaman:1984em}
\bibinfo{author}{Kaman, E.~J.}, \bibinfo{author}{Smeulders, A. W.~M.},
  \bibinfo{author}{Verbeek, P.~W.}, \bibinfo{author}{Young, I.~T.} \&
  \bibinfo{author}{Baak, J. P.~A.}
\newblock \bibinfo{journal}{\bibinfo{title}{{Image processing for mitoses in
  sections of breast cancer: A feasibility study}}}.
\newblock {\emph{\JournalTitle{Cytometry}}} \textbf{\bibinfo{volume}{5}},
  \bibinfo{pages}{244--249} (\bibinfo{year}{1984}).

\bibitem{maier2019gentle}
\bibinfo{author}{Maier, A.}, \bibinfo{author}{Syben, C.},
  \bibinfo{author}{Lasser, T.} \& \bibinfo{author}{Riess, C.}
\newblock \bibinfo{journal}{\bibinfo{title}{A gentle introduction to deep
  learning in medical image processing}}.
\newblock {\emph{\JournalTitle{Zeitschrift f{\"u}r Medizinische Physik}}}
  \textbf{\bibinfo{volume}{29}}, \bibinfo{pages}{86--101}
  (\bibinfo{year}{2019}).

\bibitem{Ciresan:2013upa}
\bibinfo{author}{Cire{\c s}an, D.~C.}, \bibinfo{author}{Giusti, A.},
  \bibinfo{author}{Gambardella, L.~M.} \& \bibinfo{author}{Schmidhuber, J.}
\newblock \bibinfo{journal}{\bibinfo{title}{{Mitosis detection in breast cancer
  histology images with deep neural networks.}}}
\newblock {\emph{\JournalTitle{MICCAI}}} \textbf{\bibinfo{volume}{16}},
  \bibinfo{pages}{411--418} (\bibinfo{year}{2013}).

\bibitem{Roux:2013kn}
\bibinfo{author}{Roux, L.} \emph{et~al.}
\newblock \bibinfo{journal}{\bibinfo{title}{{Mitosis detection in breast cancer
  histological images An ICPR 2012 contest.}}}
\newblock {\emph{\JournalTitle{J Pathol Inf}}} \textbf{\bibinfo{volume}{4}},
  \bibinfo{pages}{8} (\bibinfo{year}{2013}).

\bibitem{veta2018predicting}
\bibinfo{author}{Veta, M.} \emph{et~al.}
\newblock \bibinfo{journal}{\bibinfo{title}{Predicting breast tumor
  proliferation from whole-slide images: the tupac16 challenge}}.
\newblock {\emph{\JournalTitle{Medical image analysis}}}
  \textbf{\bibinfo{volume}{54}}, \bibinfo{pages}{111--121}
  (\bibinfo{year}{2019}).

\bibitem{He:2016ib}
\bibinfo{author}{He, K.}, \bibinfo{author}{Zhang, X.}, \bibinfo{author}{Ren,
  S.} \& \bibinfo{author}{Sun, J.}
\newblock \bibinfo{title}{{Deep Residual Learning for Image Recognition}}.
\newblock In \emph{\bibinfo{booktitle}{CVPR}}, \bibinfo{pages}{770--778}
  (\bibinfo{publisher}{IEEE}, \bibinfo{year}{2016}).

\bibitem{Li:2018ce}
\bibinfo{author}{Li, C.}, \bibinfo{author}{Wang, X.}, \bibinfo{author}{Liu, W.}
  \& \bibinfo{author}{Latecki, L.~J.}
\newblock \bibinfo{journal}{\bibinfo{title}{{DeepMitosis: Mitosis detection via
  deep detection, verification and segmentation networks}}}.
\newblock {\emph{\JournalTitle{Med Image Anal}}} \textbf{\bibinfo{volume}{45}},
  \bibinfo{pages}{121--133} (\bibinfo{year}{2018}).

\bibitem{Pati:2019kk}
\bibinfo{author}{Pati, P.}, \bibinfo{author}{Catena, R.},
  \bibinfo{author}{Goksel, O.} \& \bibinfo{author}{Gabrani, M.}
\newblock \bibinfo{title}{{A deep learning framework for context-aware mitotic
  activity estimation in whole slide images}}.
\newblock In \bibinfo{editor}{Tomaszewski, J.~E.} \& \bibinfo{editor}{Ward,
  A.~D.} (eds.) \emph{\bibinfo{booktitle}{Digital Pathology}},
  \bibinfo{pages}{7--9} (\bibinfo{publisher}{SPIE}, \bibinfo{year}{2019}).

\bibitem{scidata}
\bibinfo{author}{Bertram, C.~A.}, \bibinfo{author}{Aubreville, M.},
  \bibinfo{author}{Marzahl, C.}, \bibinfo{author}{Maier, A.} \&
  \bibinfo{author}{Klopfleisch, R.}
\newblock \bibinfo{journal}{\bibinfo{title}{A large-scale dataset for mitotic
  figure assessment on whole slide images of canine cutaneous mast cell
  tumor}}.
\newblock {\emph{\JournalTitle{Scientific Data}}} \textbf{\bibinfo{volume}{6}},
  \bibinfo{pages}{1--9} (\bibinfo{year}{2019}).

\bibitem{sliderunner}
\bibinfo{author}{Aubreville, M.}, \bibinfo{author}{Bertram, C.~A.},
  \bibinfo{author}{Klopfleisch, R.} \& \bibinfo{author}{Maier, A.}
\newblock \bibinfo{title}{{SlideRunner - A Tool for Massive Cell Annotations in
  Whole Slide Images}}.
\newblock In \emph{\bibinfo{booktitle}{Bildverarb. f{\"u}r die Med. 2018}},
  \bibinfo{pages}{309--314} (\bibinfo{publisher}{Springer},
  \bibinfo{year}{2018}).

\bibitem{Aubreville:2019um}
\bibinfo{author}{Aubreville, M.}, \bibinfo{author}{Bertram, C.~A.},
  \bibinfo{author}{Klopfleisch, R.} \& \bibinfo{author}{Maier, A.}
\newblock \bibinfo{title}{Augmented mitotic cell count using field of interest
  proposal}.
\newblock In \emph{\bibinfo{booktitle}{Bildverarbeitung f{\"u}r die Medizin
  2019}}, \bibinfo{pages}{321--326} (\bibinfo{publisher}{Springer},
  \bibinfo{year}{2019}).

\bibitem{Ren:2017kt}
\bibinfo{author}{Ren, S.}, \bibinfo{author}{He, K.}, \bibinfo{author}{Girshick,
  R.} \& \bibinfo{author}{Sun, J.}
\newblock \bibinfo{journal}{\bibinfo{title}{{Faster R-CNN: towards real-time
  object detection with region proposal networks}}}.
\newblock {\emph{\JournalTitle{IEEE Trans Pattern Anal}}}
  \bibinfo{pages}{1137--1149} (\bibinfo{year}{2017}).

\bibitem{marzahl2019deep}
\bibinfo{author}{Marzahl, C.} \emph{et~al.}
\newblock \bibinfo{journal}{\bibinfo{title}{Deep learning-based quantification
  of pulmonary hemosiderophages in cytology slides}}.
\newblock {\emph{\JournalTitle{Scientific Reports}}}
  \textbf{\bibinfo{volume}{10}}, \bibinfo{pages}{1--10} (\bibinfo{year}{2020}).

\bibitem{Lin:2017de}
\bibinfo{author}{Lin, T.-Y.}, \bibinfo{author}{Goyal, P.},
  \bibinfo{author}{Girshick, R.}, \bibinfo{author}{He, K.} \&
  \bibinfo{author}{Dollar, P.}
\newblock \bibinfo{title}{{Focal Loss for Dense Object Detection}}.
\newblock In \emph{\bibinfo{booktitle}{2017 IEEE International Conference on
  Computer Vision (ICCV)}}, \bibinfo{pages}{2999--3007}
  (\bibinfo{publisher}{IEEE}, \bibinfo{year}{2017}).

\bibitem{Ronneberger:2015gk}
\bibinfo{author}{Ronneberger, O.}, \bibinfo{author}{Fischer, P.} \&
  \bibinfo{author}{Brox, T.}
\newblock \bibinfo{title}{{U-Net - Convolutional Networks for Biomedical Image
  Segmentation.}}
\newblock In \emph{\bibinfo{booktitle}{MICCAI}}, \bibinfo{pages}{234--241}
  (\bibinfo{publisher}{Springer}, \bibinfo{year}{2015}).

\bibitem{Rahman:2016dk}
\bibinfo{author}{Rahman, M.~A.} \& \bibinfo{author}{Wang, Y.}
\newblock \bibinfo{title}{{Optimizing Intersection-Over-Union in Deep Neural
  Networks for Image Segmentation}}.
\newblock In \emph{\bibinfo{booktitle}{Advances in Visual Computing}},
  \bibinfo{pages}{234--244} (\bibinfo{publisher}{Springer, Cham},
  \bibinfo{address}{Cham}, \bibinfo{year}{2016}).

\bibitem{Veta:2014gu}
\bibinfo{author}{Veta, M.}, \bibinfo{author}{Pluim, J. P.~W.},
  \bibinfo{author}{van Diest, P.~J.} \& \bibinfo{author}{Viergever, M.~A.}
\newblock \bibinfo{journal}{\bibinfo{title}{{Breast Cancer Histopathology Image
  Analysis: A Review}}}.
\newblock {\emph{\JournalTitle{IEEE Transactions on Biomedical Engineering}}}
  \textbf{\bibinfo{volume}{61}}, \bibinfo{pages}{1400--1411}
  (\bibinfo{year}{2014}).

\bibitem{roux2014mitos}
\bibinfo{author}{Roux, L.} \emph{et~al.}
\newblock \bibinfo{journal}{\bibinfo{title}{{MITOS \& ATYPIA Detection of
  Mitosis and Evaluation of Nuclear Atypia Score in Breast Cencer Histological
  Images}}}.
\newblock {\emph{\JournalTitle{Image Pervasive Access Lab (IPAL), Agency Sci.,
  Technol. \& Res. Inst. Infocom Res., Singapore, Tech. Rep}}}
  \textbf{\bibinfo{volume}{1}}, \bibinfo{pages}{1--8} (\bibinfo{year}{2014}).

\bibitem{JKirpensteijn:2009wz}
\bibinfo{author}{Kirpensteijn, J.}, \bibinfo{author}{Kik, M.},
  \bibinfo{author}{Rutteman, G.~R.} \& \bibinfo{author}{Teske, E.}
\newblock \bibinfo{journal}{\bibinfo{title}{{Prognostic Significance of a New
  Histologic Grading System for Canine Osteosarcoma}}}.
\newblock {\emph{\JournalTitle{Vet Pathol}}} \textbf{\bibinfo{volume}{39}},
  \bibinfo{pages}{240--246} (\bibinfo{year}{2002}).

\bibitem{Loukopoulos:2007dc}
\bibinfo{author}{Loukopoulos, P.} \& \bibinfo{author}{Robinson, W.~F.}
\newblock \bibinfo{journal}{\bibinfo{title}{{Clinicopathological Relevance of
  Tumour Grading in Canine Osteosarcoma}}}.
\newblock {\emph{\JournalTitle{J Comp Pathol}}} \textbf{\bibinfo{volume}{136}},
  \bibinfo{pages}{65--73} (\bibinfo{year}{2007}).

\bibitem{Meuten:2018ir}
\bibinfo{author}{Meuten, D.}, \bibinfo{author}{Munday, J.~S.} \&
  \bibinfo{author}{Hauck, M.}
\newblock \bibinfo{journal}{\bibinfo{title}{{Time to Standardize? Time to
  Validate?}}}
\newblock {\emph{\JournalTitle{Vet Pathol}}} \textbf{\bibinfo{volume}{55}},
  \bibinfo{pages}{195--199} (\bibinfo{year}{2018}).

\end{thebibliography}

\end{document}